\documentclass{article}

\usepackage{graphicx}
\usepackage{amssymb}
\usepackage{amstext}
\usepackage{epsfig}
\usepackage{fullpage}
\usepackage{caption}
\usepackage{subcaption}

\usepackage{array,multirow}

\usepackage{latexsym}
\usepackage{amsmath}
\usepackage{bm} 
\usepackage{bbm}

\usepackage{url}

\usepackage{graphicx}
\usepackage{amssymb}
\usepackage{amstext}
\usepackage{epsfig}

\usepackage{latexsym}
\usepackage{amsmath}
\usepackage{bm} 
\usepackage{bbm}

\usepackage{array,multirow}

\usepackage{amsthm}
\usepackage{amsmath}
\usepackage{amsfonts}
\usepackage{amssymb,amstext}
\usepackage{appendix}
\usepackage{enumitem}

\usepackage{algorithm}
\usepackage[noend]{algpseudocode}
\makeatletter
\def\BState{\State\hskip-\ALG@thistlm}
\makeatother

\usepackage{epsfig}
\usepackage{algpseudocode}
\usepackage{amscd}
\usepackage{algorithm}
\usepackage{tikz}

\newcommand{\beq}{\begin{equation}}
\newcommand{\eeq}{\end{equation}}

\newcommand{\beqa}{\begin{eqnarray}}
\newcommand{\eeqa}{\end{eqnarray}}

\newcommand{\bal}{\begin{align}}
\newcommand{\eal}{\end{align}}

\newcommand{\bsp}{\begin{equation}\begin{split}}
\newcommand{\esp}{\end{split}\end{equation}}

\newcommand{\bit}{\begin{itemize}}
\newcommand{\eit}{\end{itemize}}

\newcommand{\ben}{\begin{enumerate}}
\newcommand{\een}{\end{enumerate}}

\newcommand{\AR}{\mathbb{R}}

\title{Expressive recommender systems through normalized nonnegative models}

\author{Cyril J.~Stark\footnote{Massachusetts Institute of Technology, 77 Massachusetts Avenue, 6-304, Cambridge MA 02139-4307, USA}}

\date{\today}

\begin{document}

\maketitle
\begin{abstract}

We introduce normalized nonnegative models (NNM) for explorative data analysis. NNMs are partial convexifications of models from probability theory. We demonstrate their value at the example of item recommendation. We show that NNM-based recommender systems satisfy three criteria that all recommender systems should ideally satisfy:  high predictive power, computational tractability, and expressive representations of users and items. Expressive user and item representations are important in practice to succinctly summarize the pool of customers and the pool of items. In NNMs, user representations are expressive because each user's preference can be regarded as normalized mixture of preferences of stereotypical users. The interpretability of item and user representations allow us to arrange properties of items (e.g., genres of movies or topics of documents) or users (e.g., personality traits) hierarchically. 

%

\end{abstract}

\section{Introduction}

Recommender systems are algorithms designed to recommend items to users. Good recommender systems address the following three partially conflicting objectives: (1) \emph{predictive power} (despite very sparse and noisy data), (2) \emph{computational tractability} (despite quickly growing numbers of users and items), (3) \emph{interpretability} (to allow for feedback, market analysis and visual representations). 
Here, to address those difficulties, we are going to adopt the \emph{system-state-measurement} paradigm  in the form of a class of models which we call \emph{normalized nonnegative models} (NNM). 

Adopting the system-state-measurement paradigm amounts to making a clear distinction between the `state of a system' and the `measurement device' used to probe that system. The success of this paradigm in science and engineering motivates its application in item recommendation and beyond. In the study of recommendation the \emph{system} is that abstract part of our thinking that determines whether we like or dislike an item. The \emph{state} of that system varies from person to person; it forms the description of the individual preferences. The \emph{measurements} that we perform on the system are questions of the form ``Do you like the movie \emph{Jurassic Park}?". Each measurement probes our taste (e.g., movie taste) and measuring sufficiently many diverse questions allows to get an idea of the preferences/opinion of a person. In the natural sciences and engineering the system is oftentimes described in terms a \emph{sample space}, the state of the system is \emph{probability distribution} on that sample space and a measurement is a \emph{random variable}. Hence, ideally, to adopt that picture, we need to compute the following building blocks: (1) an effective sample space, (2) a probability distribution for each user to describe that user's taste, (3)  a random variable for each item to describe questions like ``How do you rate the movie \emph{Ex Machina}?". To arrive at NNMs we simply convexify the third of these building blocks, i.e., the space of random variables. This convex relaxation will allow us to compute NNMs through alternating convex optimization. In this manner, approximate inference of NNMs becomes \emph{computationally tractable}. 

The main strength of NNMs are highly interpretable user and item representations. The way we represent users allows us to regard users as normalized mixtures of a small number of \emph{user stereotypes}. We provide strategies to characterize those stereotypical users in words so that those stereotypes can be understood intuitively by people who are unfamiliar with data analysis. Hence, NNMs allow everybody to interpret users' behaviors as mixtures of well-characterized stereotypical behaviors. On the other hand, the way we represent items allows us to infer hierarchical orderings of item categories like movie genres or topics of documents. This is how we address the criterion \emph{interpretability}. Of course, topic models like those based on latent Dirichlet allocation (LDA) have also been used to derive said expressive descriptions of users and items. We explain this in much more detail when discussing related work towards the end of this paper.

Finally, we evaluate the last remaining criterion \emph{predictive power} in numerical experiments. We show that in mean-average-error, NNMs outperform methods like SVD++~\cite{koren2008factorization} on MovieLens datasets. This indicates that the high level of interpretability of NNMs \emph{comes not at the price of sacrificing predictive power}.

Throughout the paper we introduce NNMs through their application in item recommendation. But we hope that our presentation will be clear enough to convince the reader that the scope of NNMs is not limited to recommendation---in the very same sense that the scope of probability theory is not limited to one particular branch of science.


\section{Notation}

For $n \in \mathbb{N}$, we set $[n] = \{1,...,n\}$. Throughout, $u \in [U]$ labels users, $i \in [I]$ labels items and $z \in [Z]$ denote possible ratings (e.g., $z \in [5]$ in case of 5-star ratings). By $R \in [Z]^{U \times I}$ we denote the complete rating matrix, i.e., $R_{ui} \in [Z]$ is the rating that user $u$ provides for item $i$. In practice, we only know a subset of the entries of $R$. We use $\Gamma \subseteq [U] \times [I]$ to mark the known entries, i.e., $(u,i) \in \Gamma$ if $R_{ui}$ is known a priori. We use $\Delta = \{ \vec{p} \in \AR^D_+ | \| \vec{p} \|_1 = 1 \}$ to denote the probability simplex. A finite probability space is described in terms of a sample space $\Omega = \{ \omega_1, ..., \omega_D \}$, probability distributions $\vec{p} \in \Delta$ and random variables $\hat{E}$ with some alphabet $[Z]$. Recall that random variables are functions $\hat{E} : \Omega \rightarrow [Z]$.


\section{Probability theory recap}\label{Sect:prob.theory.recap}

According to Kolmogorov, a random experiment with finite sample space is described by the following triple:
\begin{itemize}
\item		A sample space $\Omega = \{ \omega_1, ..., \omega_D\}$. The elements $\omega_j$ denote elementary events.
\item		A probability distribution $\vec{p} \in \mathbb{R}^D_+$ with $\sum_j (\vec{p})_j = 1$, i.e., $\vec{p}$ is an element of the probability simplex $\Delta$.
\item		A random variable $\hat{E}$, i.e., a function $\hat{E}: \Omega \rightarrow \{ 1,...,Z \}$ for some alphabet size $Z \in \mathbb{N}$.
\end{itemize}
We denote by $\mathbb{P}[ \hat{E} = z ]$ the probability of the event $\{ \omega \in \Omega | \hat{E}(\omega) = z \}$. Therefore, 
\beq\label{fwfgergrdd}
	\mathbb{P}[ \hat{E} = z ] = \mathbb{P}[ \hat{E}^{-1}(z) ] = \sum_{\omega \in \hat{E}^{-1}(z)} p_{\omega}
\eeq
where $\hat{E}^{-1}(z) \subseteq \Omega$ denotes the pre-image of $z$ under the map $\hat{E}$. The expression~\eqref{fwfgergrdd} can be rewritten using indicator vectors. For that purpose we define $\vec{E}_z$ by
\beq\label{fej435hjfwefehj}
	\bigl( \vec{E}_{z} \bigr)_{j}
	= \left\{ \begin{array}{ll}  1,   & \text{ if $\omega_j \in \hat{E}^{-1}(z)$}      \\ 0,  &\text{ otherwise.}   \end{array} \right.
\eeq
It follows that $\mathbb{P}[ \hat{E} = z ] = \vec{E}_z^T \vec{p}$
,i.e., probabilities for measuring specific outcomes of random variables can be expressed in terms of inner products between two $D$-dimensional vectors. By~\eqref{fej435hjfwefehj}, $\sum_z \vec{E}_z = (1,...,1)^T$.

\emph{Examples.} For an unbiased coin, $\vec{p} = (1/2,1/2)^T$, $\vec{E}_{\text{heads}} = (1,0)^T$ and $\vec{E}_{\text{tails}} = (0,1)^T$. Consequently, $\mathbb{P}[\hat{E} = \text{heads}] = (1/2,1/2)(1,0)^T = 1/2$. For a biased 4-sided coin, we may have $\vec{p} = (1/4,1/4,1/8,3/8)^T$ and $\vec{E}_z$ such that $(\vec{E}_z)_j := \delta_{zj}$. It follows for example that $\mathbb{P}[\hat{E} = 1 \text{ or } 4] = \vec{p}^T (\vec{E}_1 + \vec{E}_4) = 5/8$.

\section{Normalized nonnegative models}\label{Sect:Normalized.nonnegative.models}

We adopt the system-state-measurement paradigm. The \emph{system} we are interested in is the part of our mind that determines the outcome to the question ``Do you like item $i$?" ($i \in [I]$). For each user $u \in [U]$ this system is in some state described by a distribution $\vec{p}_u \in \Delta$ on some unknown sample space $\Omega = \{ \omega_1, ..., \omega_D \}$ representing the system. Each question ``Do you like item $i$?" is modeled in terms of a random variable $\hat{E}_i$ with alphabet $[Z]$ ($Z=5$ for 5-star ratings). We denote by $\mathbb{P}_u[ \hat{E}_i = z ]$ the probability for user $u$ to rate item $i$ with $z \in [Z]$. Thus, 
\beq\label{few45jkljfnk}
	\mathbb{P}_u[ \hat{E}_i = z ] = \mathbb{P}_u[ \hat{E}^{-1}_i(z) ] = \vec{E}_{iz}^T \vec{p}_u
\eeq
where $\vec{p}_u$ models the state of user $u$ and where $\vec{E}_{iz} \in \{0,1\}^D$ is defined by
\beq\label{fej435hjhj}
	\bigl( \vec{E}_{iz} \bigr)_j 
	= \left\{ \begin{array}{ll}  1,   & \text{ if $\omega_j \in \hat{E}^{-1}_i(z)$}      \\ 0,  &\text{ otherwise.}   \end{array} \right.
\eeq
By~\eqref{fej435hjhj}, $\sum_z \vec{E}_{iz} = (1,...,1)^T$. We denote by $\mathcal{M}_0$ the set of all valid descriptions $( \vec{E}_{1}, ..., \vec{E}_{Z} )$ of a random variable $\hat{E}$, i.e., 
\[
	\mathcal{M}_0 = \Bigl\{ ( \vec{E}_{1}, ..., \vec{E}_{Z} ) \in \{0,1\}^{D \times Z } \Bigl|  \sum_z \vec{E}_{iz} = (1,...,1)^T \Bigr\}.
\] 
Allowing for stochastic mixtures of elements of $\mathcal{M}_0$ we arrive at the \emph{convex relaxation} 
\[
	\mathcal{M} = \Bigl\{ ( \vec{E}_{1}, ..., \vec{E}_{Z} ) \in \AR_+^{D \times Z } \Bigl|  \sum_z \vec{E}_{iz} = (1,...,1)^T \Bigr\}
\] 
of $\mathcal{M}_0$. In the remainder, the tuple of vectors $\bigl( (\vec{p}_u)_{u \in [U]}, (\vec{E}_{iz})_{i \in [I], z \in [Z]} \bigr)$ denotes a \emph{normalized nonnegative model} (NNM) if $\vec{p}_u \in \Delta$ for all users $u \in [U]$ and if $(\vec{E}_{iz})_{z \in [Z]} \in \mathcal{M}$ for all items $i \in [I]$. A NNM captures observations well if $R \approx \left( \mathrm{argmax}_z \{ \vec{E}_{iz}^T \vec{p}_u \}_{z \in [Z]} \right)_{u \in [U], i \in [I]}$.

\subsection{Non-categorical}

In the previous description of the random variables $\hat{E}_i$ we did not make any assumptions on the nature of the outcomes $z \in [Z]$. Hence, the outcomes $z \in [Z]$ are \emph{categorical}, i.e., the outcomes are not ordered (e.g., $(z=3) < (z=4)$) and the numerical values $z \in [Z]$ carry no meaning. This is a feature of NNMs as it makes them applicable in a very wide range of settings. 

In item recommendation, systems that allow for categorical information are particularly convenient to make use of side information such as the gender of users. But even if we only focus on the user-item matrix, systems that allow for categorical information may generally have an advantage over other systems because 5-star ratings do not come with a scale. For example, we cannot claim that we prefer a 2-star-item over a 1-star-item to the same extent as we prefer a 3-star-item over a 2-star-item. 

On the other hand, star-ratings are clearly ordered as, for example, a 4-star-rating is better than a 3-star-rating. Consequently, we lose information if we treat ratings in a purely categorical manner. Therefore, to avoid the little-data-problem we may want to interpret a rating $z \in [Z]$ of an item $i$ by user $u$ as an approximation of $\mathbb{P}[u \text{ likes } i]$, i.e., 
\beq\label{Eq:alternative.interpretation.of.data}
	z/Z \approx \mathbb{P}[u \text{ likes } i].
\eeq
To model this interpretation of users' ratings, we regard the random variables $\hat{E}_i$ ($i \in [I]$) as binary random variables whose outcomes are interpreted as `\emph{I like item $i$}' and `\emph{I dislike item $i$}', respectively.

NNMs allow for the modeling of categorical variables because we assigned individual vectors $(\vec{E}_{iz})_{z \in [Z]}$ to each of the outcomes $z \in [Z]$. This can be done in general matrix factorization (e.g., SVD++~\cite{koren2008factorization}), and the potential possibilities motivate a thorough analysis of the modeling of categorical variables in terms of general matrix factorization.

%

\section{Interpretability}\label{sect:NNM.interpretability}

One of the reasons for the popularity of models from probability theory is their interpretability. To make this more precise, we imagine flipping a coin. The natural probabilistic description of a single coin flip is in terms of a sample space $\Omega = \{ \omega_{\text{heads}}, \omega_{\text{tails}} \}$ where $\omega_{\text{heads}}$ is the event \emph{heads} and $\omega_{\text{tails}}$ is the event \emph{tails}. We denote by $\vec{\delta}_{\text{heads}}$ and $\vec{\delta}_{\text{tails}}$ the deterministic distributions located at $\omega_{\text{heads}}$ and $\omega_{\text{tails}}$, respectively. (For example, $\bigl( \vec{\delta}_{\text{heads}} \bigr)_{\omega} = 1$ if $\omega = \omega_{\text{heads}}$ and $\bigl( \vec{\delta}_{\text{heads}} \bigr)_{\omega} = 0 $ otherwise.) Hence, we can easily and clearly describe the states $\vec{\delta}_{\text{heads}}$ and $\vec{\delta}_{\text{tails}}$ in terms of words: $\vec{\delta}_{\text{heads}}$ is the state that always returns \emph{heads}, and $\vec{\delta}_{\text{tails}}$ is the state that always returns \emph{tails}. It is through those concise descriptions that we get an \emph{intuitive understanding} of the states $\vec{\delta}_{\text{heads}}$ and $\vec{\delta}_{\text{tails}}$.

Every possible state $\vec{p} = (p_{\text{heads}},p_{\text{tails}})$ of the coin is a probabilistic mixture of $\vec{\delta}_{\text{heads}}$ and $\vec{\delta}_{\text{tails}}$. Since we have an intuitive understanding of what mixtures are, we can thus lift our intuitive understanding of $\vec{\delta}_{\text{heads}}$ and $\vec{\delta}_{\text{tails}}$ to arrive at an intuitive understanding of general states $\vec{p} = (p_{\text{heads}},p_{\text{tails}})$. We think that this is one of the main strengths of probabilistic models and we think that this is one of the main reasons why those models are so appealing not only to scientists and engineers but also to people with less mathematical training.

In NNMs the states of users are distributions $\vec{p}_u = \sum_{\omega=1}^D p_{u, \omega}  \vec{\delta}_{\omega}$ where $(\vec{\delta}_{\omega})_k = 1$ if $\omega=k$ and $(\vec{\delta}_{\omega})_k = 0$ otherwise. Hence, \emph{we could get a good understanding of the users' states if we found ways to describe the elementary preferences $\vec{\delta}_{\omega}$ in an intuitive manner}. We call those elementary preferences $\vec{\delta}_{\omega}$ \emph{stereotypes} because each user's preference is a mixture of those stereotypes. We next describe approaches to acquire said intuitive description of stereotypes.

\subsection{Understanding stereotypes through tags}

Oftentimes, we not only have access to users' ratings of items but we also have access to side information about items in terms of tags. For example the MovieLens 1M dataset provides genre tags for movies; each movie gets assigned to (sometimes multiple) genres like \emph{Action, Adventure, Animation,} etc. Note that we do have an intuitive understanding of those genres---just as we have an intuitive understanding of the coin-states $\vec{\delta}_{\text{heads}}$ and $\vec{\delta}_{\text{tails}}$. Thus, at the example of genre tags we explain next, how side information can be used to get an intuitive characterization of stereotypes.

We assume that each movie $i$ is assigned to some genres $g^{i}_{1}, ..., g^{i}_{n_i} \in \{ g_1, ..., g_G \}$. To characterize a stereotype $\vec{\delta}_{\omega}$ we want to determine how much the hypothetical user $\vec{\delta}_{\omega}$ likes movies of genre $g_1$, movies of genre $g_2$, etc. We make this precise in terms of the following game to characterize stereotypes $\vec{\delta}_{\omega}$. The game involves a referee and two players Alice and Bob. For some $\omega \in \Omega$, Alice's user vector is assumed to be $\vec{\delta}_{\omega}$. We proceed as follows.


\begin{enumerate}
\item Fix a genre $g \in \{ g_1, ..., g_G \}$. Let $i_1, ..., i_m$ denote all the movies that have been tagged with $g$.
\item Bob is given access to Alice's vector $\vec{\delta}_{\omega}$, to $g$, to all genre tags $\bigl( g^{i}_{1}, ..., g^{i}_{n_i} \bigr)_{i \in [I]}$ and to all item vectors $\bigl( \vec{E}_{iz} \bigr)_{i \in [I], z \in [Z]}$ ($z \in [2]$; $z =1$ means `like' and $z=2$ means `dislike').
\item The referee draws uniformly at random a movie $i^*$ from $\{ i_1, ..., i_m \}$. 
\item By looking up $R$, the referee checks whether Alice likes or dislikes $i^*$. We denote her answer by $z^* \in \{\text{like}, \text{dislike}\}$.
\item Bob guesses $z^*$. He wins the game if he guesses $z^*$ correctly. Otherwise, he loses.
\end{enumerate}

Before we describe how to make use of this game for the characterization of stereotypical users, we describe Bob's strategy. First we note that Bob needs to estimate the probability for $z^* = 1$ and $z^* = 2$, respectively. Conditioned on the event `\emph{Referee draws $i$}' we have that $\mathbb{P}[ z^* = 1 \; | \; i ] = \vec{E}_{i1}^T \vec{p}$. The probability that the referee draws $i$ is $1/m$ because there are in total $m$ movies associated with $g$. Therefore,
\[
	\mathbb{P}[ z^* = 1 ] = \sum_{i \in \{ i_1, ..., i_m \}} \mathbb{P}[ z^* = 1 \; | \; i ] \; \mathbb{P}[i] = \vec{E}_g^T \vec{p}.
\]
where
\[
	\vec{E}_g := \frac{1}{m} \sum_{i \in \{ i_1, ..., i_m \}} \vec{E}_{i1}.
\]
Hence, Bob computes $\vec{E}_g^T \vec{\delta}_{\omega} = E_{g,\omega}$. If $E_{g,\omega} \geq 1/2$ he guesses $z^* = 1$. Otherwise, he guesses $z^* = 2$.

How can this game be used for the characterization of stereotypes? The number $E_{g,\omega}$ specifies the probability that the stereotypical user $\vec{\delta}_{\omega}$ likes a \emph{random} movie from genre $g$. For instance, if $E_{g,\omega} \approx 1$ for $g \in [G]$ then we know that the stereotypical user $\vec{\delta}_{\omega}$ very much likes movies from genre $g$. We repeat above game for all $g \in \{ g_1, ..., g_G \}$ and for all $\omega \in \Omega$. We arrive at numbers $( E_{g,\omega} )_{g \in [G],\omega \in \Omega}$. For each $\omega$, the tuple $( E_{g,\omega} )_{g \in[G]}$ provides a characterization of the preferences of the stereotypical user $\vec{\delta}_{\omega}$. The characterization $( E_{g,\omega} )_{g \in[G]}$ is convenient because $E_{g,\omega}$ specifies the probability for $\vec{\delta}_{\omega}$ to like a movie from genre $g$, and because those genres $g$ are understood intuitively.

\subsection{Understanding stereotypes without tags}

In the previous section we proposed a method for characterizing stereotypes $\vec{\delta}_{\omega}$. That method is applicable whenever items come along with interpretable tags. What can we do if no such tags are available? Assume we only have access to users' ratings of items. In those cases we suggest to proceed by characterizing each of the stereotypical users through a list of items they like. For a stereotype $\vec{\delta}_{\omega}$, those items can be found by firstly, collecting all items with the property that $\bigl( \vec{E}_{iZ} \bigr)_{\omega} \approx 1$ (e.g., vectors associated to 5 stars). Denote those highly rated movies by $\gamma^{\omega}_{1}, ..., \gamma^{\omega}_{M}$. Then, in a second step, we select from the set $\{ \gamma^{\omega}_{1}, ..., \gamma^{\omega}_{M} \}$ those items that are popular (i.e., known to many people). We arrive at items $\gamma^{\omega}_{m_1}, ..., \gamma^{\omega}_{m_J}$. To characterize the stereotype $\vec{\delta}_{\omega}$, we report the list $\gamma^{\omega}_{m_1}, ..., \gamma^{\omega}_{m_J}$.

\subsection{Stereotypes in general matrix factorization models}

Let $\vec{r}_u$ denote user vectors computed in a general matrix factorization model. Computing the convex hull of the cone spanned by the set $\{ \vec{r}_u \}_{u \in [U]}$  we could in principle determine user vectors $\vec{r}_{i_1}, ..., \vec{r}_{i_T}$ with the property that for all $\vec{r}_u$ there exist coefficients $\lambda_1, ..., \lambda_T \geq 0$ such that $\sum_{k=1}^T \lambda_k \vec{r}_{i_k} = \vec{r}_u$. Therefore, as in NNMs, we can still express every user vector as mixture of other user vectors. 

There are, however, at least two major problems with this approach. Firstly, computing the convex conic hull of the span of $\{ \vec{r}_u \}_{u \in [U]}$ is computationally not tractable; even for small numbers of users. Secondly, we expect the number $T$ of extremal rays $\AR \vec{r}_{i_1}, ..., \AR  \vec{r}_{i_T}$ of the convex conic hull to be very large---independently of $D$. Therefore, users' states are difficult to interpret because they are the mixture of a very large number of stereotypes. We would like to stress that for each dimension $D$, NNMs are efficient as they only use the least possible number of $D$ stereotypes. For example, if we only needed $D-1$ stereotypes then all vectors could be restricted to a sub-space of $\AR^D$ and a ($D-1$)-dimensional NNM could be used instead of the $D$-dimensional NNM.


%

\section{Hierarchical structure of tags}\label{sect:NNM.inference.of.hierarchy.of.tags}

Assume the considered items are tagged. For instance, as before, if the items are movies then these tags could specify which genre each movie belongs to. We denote by $\{ i^t_1, ..., i^t_{m_t} \}$ all items that have been tagged with a tag $\tau_t$ from the set of all tags $\{ \tau_t \}_{t \in [T]}$. As before, we describe tags $\tau_t$ ($t \in [T]$) in terms of vectors
\beq\label{gekngjkegjkrgk}
	\vec{E}_{\tau_t} := \frac{1}{m_t} \sum_{i \in \{ i_1, ..., i_{m_t} \}} \vec{E}_{iZ},
\eeq
so that $\vec{p}_u^T \vec{E}_{\tau_t}$ is the probability that user $u$ likes a randomly chosen item $i \in \{ i^t_1, ..., i^t_{m_t} \}$. As we are going to see next,~\eqref{gekngjkegjkrgk} enables us to order tags in a hierarchical manner. 

We note that if the tag vectors $\vec{E}_{\tau_t}$ were binary vectors (i.e., $ \in \{ 0,1 \}^D$), then we would say $\tau_t \subseteq \tau_{t'}$ whenever the support of $\vec{E}_{\tau_t}$ is contained in the support of $\vec{E}_{\tau_{t'}}$. This is a meaningful definition of `$\subseteq$' for tags because if $\tau_t \subseteq \tau_{t'}$ then \emph{`$\vec{\delta}_{\omega}$ likes $\tau_t$'} implies \emph{`$\vec{\delta}_{\omega}$ likes $\tau_{t'}$'}.

Generally, in NNMs, tag vectors $\vec{E}_{\tau_t}$ are not binary and therefore, the definition of $\tau_t \subseteq \tau_{t'}$ needs to make sense for non-binary vectors. To find a new definition of `$\subseteq$' we note that in the previous binary setting, $\tau_t \subseteq \tau_{t'}$ if and only if
\beq\label{gewgerg}
	\vec{E}_{\tau_{t'}}^T \vec{E}_{\tau_{t}} =   \sum_{\omega \in \Omega}  \bigl( \vec{E}_{\tau_{t}} \bigr)_{\omega} =: \| \vec{E}_{\tau_{t}} \|_1.
\eeq
Condition~\eqref{gewgerg} can never be satisfied if the components of $\vec{E}_{\tau_t}$ and $\vec{E}_{\tau_{t'}}$ are $< 1$. This can be the case in NNMs. However, the operational meaning of the condition~\eqref{gewgerg} is preserved under the relaxation
\beq\label{gewger235g}
	\vec{E}_{\tau_{t'}}^T \vec{E}_{\tau_{t}} \geq (1 - \varepsilon) \| \vec{E}_{\tau_{t}} \|_1
\eeq
if $\varepsilon > 0$ is small. That is because if the relaxed condition~\eqref{gewger235g} is satisfied then we still have that most of the weight of $\vec{E}_{\tau_t}$ is contained in the approximate support of $\vec{E}_{\tau_{t'}}$. 

Therefore, we say $\tau_t \subseteq_{\varepsilon} \tau_{t'}$ if condition~\eqref{gewger235g} is satisfied, and we say $\tau_t =_{\varepsilon} \tau_{t'}$ if both $\tau_t \subseteq_{\varepsilon} \tau_{t'}$ and $ \tau_{t'} \supseteq_{\varepsilon} \tau_t$. 

The collection of all relations `$\subseteq_{\varepsilon}$' between tags can be represented graphically in terms of a graph. For that purpose we interpret the set of tags $\{ \tau_t \}_{t \in [T]}$ as the vertex set $V$ of a graph $G = (V,E)$. $G$ contains directed edges defined through the rule
\[
	( \tau_{t} \rightarrow \tau_{t'} ) \ \text{ if } \tau_{t'} \subseteq_{\varepsilon} \tau_{t}.
\]
For every choice of $\varepsilon \in [0,1]$, the graph $G$ induces an approximate hierarchical ordering of tags; see figure~\ref{fig:ML_1M_hierarchy}.

\section{Computation of normalized nonnegative models}

Presumably, the simplest approach for computing a NNM proceeds via alternating constrained optimization to solve
\beq\label{fejw4h5jh}
	\min_{\vec{p}_u \in \Delta, (\vec{E}_{iz})_{z \in [Z]} \in \mathcal{M}} \sum_{(u,i) \in \Gamma} \bigl( \vec{E}_{iz}^T \vec{p}_u - R_{ui}/Z \bigr)^2,
\eeq
i.e., the algorithm switches back and forth between optimizing $(\vec{p}_u)_{u \in [U]}$ (keeping $(\vec{E}_{iz})_{i \in [I], z\in [Z]}$ fixed) and optimizing $(\vec{E}_{iz})_{i \in [I],z \in [Z]}$ (keeping $(\vec{p}_u)_{u \in [U]}$ fixed). Each of these tasks can be computed efficiently and in parallel. Moreover, this approach is guaranteed to converge to a local minimum. 

In the context of recommendation, training data is typically subject to a large \emph{selection bias}: a majority of the ratings are high ratings. This significantly impacts the model we fit to the data. For example, if all the known entries (marked by $\Gamma$) of the rating matrix $R$ were equal to $Z$ then a 1-dimensional model (user vectors $=1$, item vectors $= Z$) would lead to zero training error. It is commonly believed that $(u,i) \in \Gamma^c$ implies typically that $u$ does not like $i$. Thus, during the first 2 iterations, we set the missing entries of $R$ equal to zero. We believe that this preprocessing leads to an initialization for alternating constrained optimization that captures more accurately the ground truth. We arrive at Algorithm~\ref{Alg:Constrained.least.squares.for.NNM} to fit NNMs to measured data. 

\begin{algorithm}
\caption{Alternating constrained least squares for NNM}\label{Alg:Constrained.least.squares.for.NNM}
\begin{algorithmic}[1]
\State 	Fix $D$ (e.g., by cross validation).
\State	For all $u$, initialize $\vec{p}_u = \vec{e}_{i}$ where $i \in [D]$ is chosen uniformly at random and where $(\vec{e}_i)_j = \delta_{ij}$.
\State	For all items $i$, solve the (linearly constrained) nonnegative least squares problem $ \label{efhjeh}\min_{(\vec{E}_{iz})_{z \in [Z]} \in \mathcal{M}} \sum_{u: (u,i)\in\Gamma} \bigl( \vec{E}_{iz}^T \vec{p}_u - R_{ui}/Z \bigr)^2$.
\State	For all users $u$, solve the (linearly constrained) nonnegative least squares problem $\min_{\vec{p}_u \in \Delta} \sum_{i: (u,i)\in\Gamma} \bigl( \vec{E}_{iz}^T \vec{p}_u - R_{ui}/Z \bigr)^2$.\label{43j5hjhfe} \label{fwk4k5j345jn}
\State 	Repeat steps 3 and 4 until a stopping criteria is satisfied; e.g., until a maximum number of iterations is reached. \emph{For the first 2 iterations} we pretend we knew all of $R$ by setting unknown entries equal to zero.
\end{algorithmic}
\end{algorithm}

\section{Computational tractability}

All of the steps in Algorithm~\ref{Alg:Constrained.least.squares.for.NNM} can be parallelized straightforwardly. The main bottleneck are large number of summands in the objective functions of user and item updates. With standard methods this becomes a potential issue during the first two iterations (i.e., preprocessing). There are at least two loopholes. The easiest solution is in terms of sampling a fixed number of unknown entries and replacing only those with zeros. Here, the number of sampled entries should be comparable to the number of known entries so that we can compensate for the selection bias towards positive ratings. 

Alternatively, we can run Algorithm~\ref{Alg:Constrained.least.squares.for.NNM} for a subset of all users and items. We denote these users and items by $\{ u_n \}_{n=1}^N$ and $\{ i_m \}_{m=1}^M$, respectively. If $N$ and $M$ equal a couple of thousands then we can easily run Algorithm~\ref{Alg:Constrained.least.squares.for.NNM}; see section `Experiments'. How can we compute representations of the remaining users and items? To determine user vectors for $u \notin \{ u_n \}_{n=1}^N$ we simply solve step~\ref{fwk4k5j345jn} of Algorithm~\ref{Alg:Constrained.least.squares.for.NNM} to determine $\vec{p}_u$. We can proceed analogously to determine representations for $i \notin \{ i_m \}_{m=1}^M$. These single small optimization problems can be solved quickly and therefore, online. The resulting scheme is a two-phase procedure where we compute the representations of the `anchor users' $\{ u_n \}_{n=1}^N$ and `anchor items' $\{ i_m \}_{m=1}^M$ offline, and where we compute or update all other user and item representations online. Of course, this requires selecting the sets $\{ u_n \}_{n=1}^N$ and $\{ i_m \}_{m=1}^M$ so that the anchor items are popular ($\rightarrow u$ knows some of them), and so that the anchor users rate lots of items ($\rightarrow$ soon after release, $i$ is rated by a couple of those users).

\section{Experiments}\label{Sect:experiments}


We evaluate predictive power and interpretability of NNMs at the example of the omnipresent MovieLens (ML) 100K and 1M datasets.\footnote{http://grouplens.org/datasets/movielens/} Due to lack of space we moved the details about the configurations of all algorithms and  results for the MovieLens 100K dataset to the appendix. To use the data economically, we employed interpretation~\eqref{Eq:alternative.interpretation.of.data} of the rating matrix $R$. We computed NNMs in Matlab using cvx~\cite{cvx} calling SDPT3~\cite{toh1999sdpt3}.

\begin{figure}
\centering
\begin{subfigure}{.5\textwidth}
  \centering
  \includegraphics[width=1\linewidth]{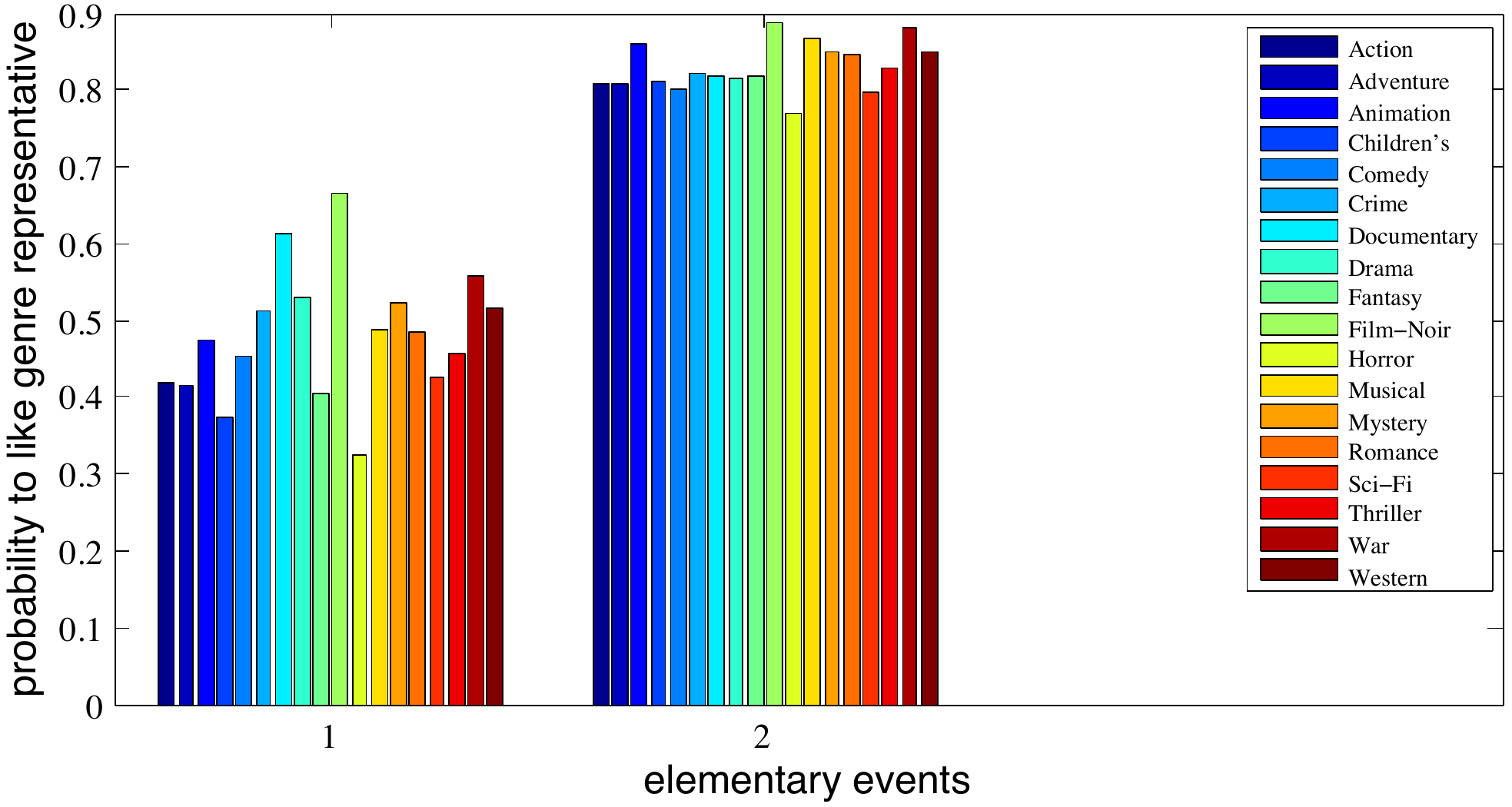}
\end{subfigure}%
\begin{subfigure}{.5\textwidth}
  \centering
  \includegraphics[width=0.96\linewidth]{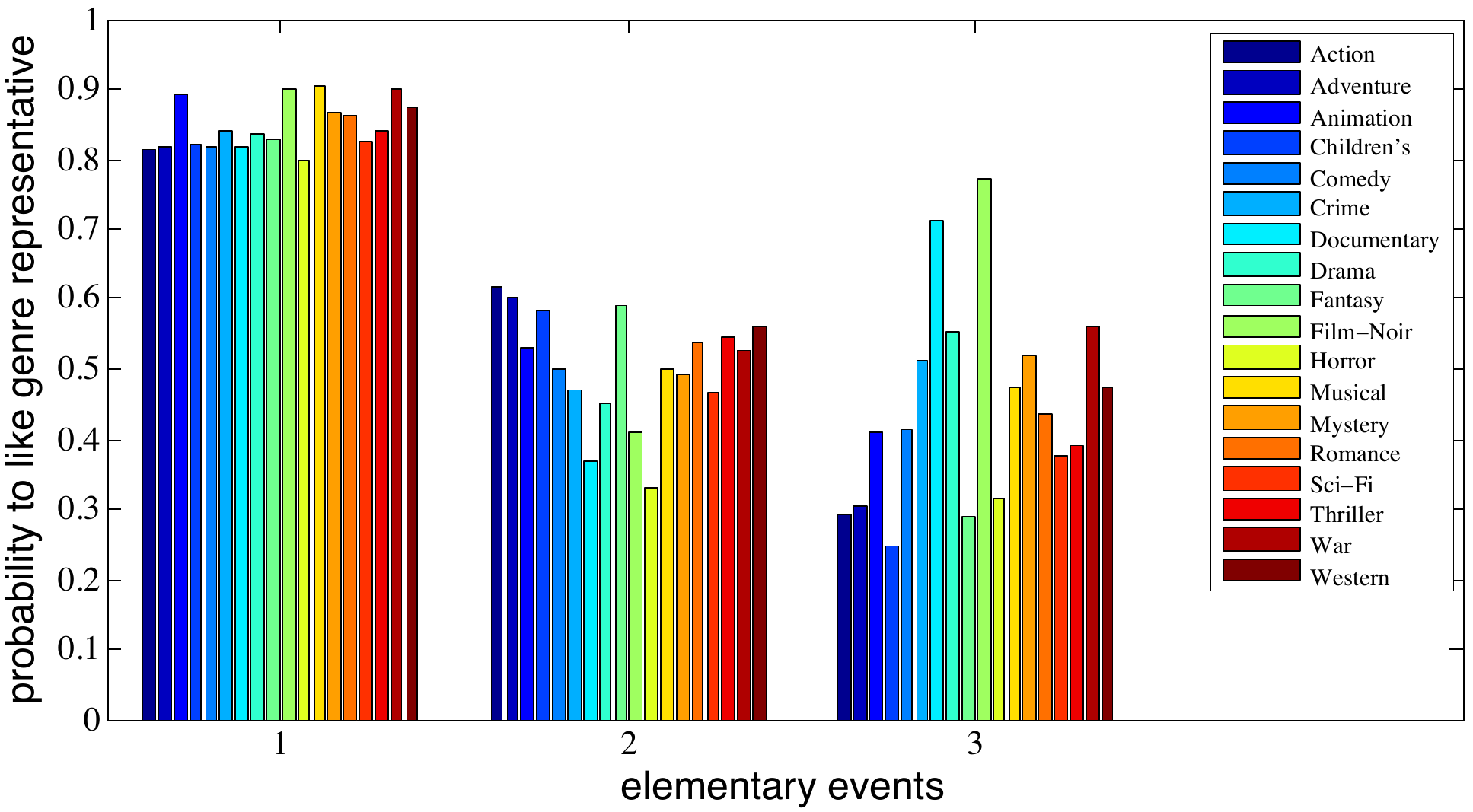}
\end{subfigure}
\caption{Left: Genre profiles for stereotype at $D = 2$; ML 1M. Right: Genre profiles for stereotype at $D = 3$; ML 1M.}
\label{fig:ML_1M_genre_profiles_of_stereotypes}
\end{figure}

\begin{figure*}[tbp]
\centering
\includegraphics[width=0.8\columnwidth]{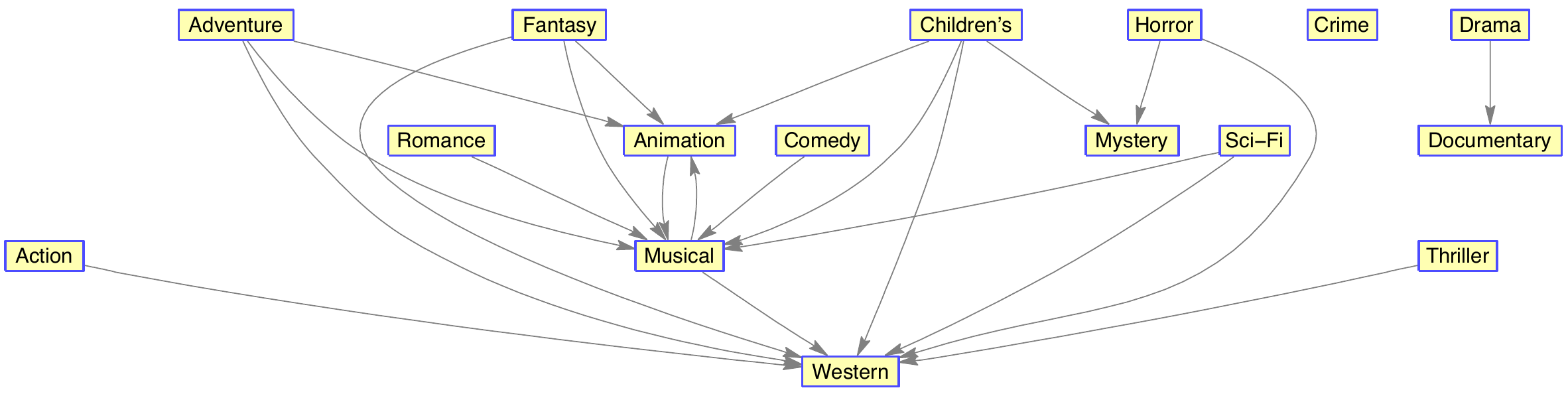}
\caption{Genre hierarchy at $\varepsilon = 1/3$ and $D = 8$; extracted from MovieLens 1M dataset.}
\label{fig:ML_1M_hierarchy}
\end{figure*}

\emph{Interpretability.} We computed figure~\ref{fig:ML_1M_genre_profiles_of_stereotypes} to illustrate the interpretability of NNMs through user stereotypes. We computed figure~\ref{fig:ML_1M_hierarchy} to evaluate interpretability of NNMs through emergent hierarchies of tags.

Figure~\ref{fig:ML_1M_genre_profiles_of_stereotypes} (left) corresponds to a 2-dimensional NNM. The two towers characterize the two user stereotypes. The second stereotype does not much care about the particular genre; she rates many movies highly and is open to everything. The first stereotype is generally more skeptical but rates movies from the genres `Documentaries' and `Film-Noir' highly; she dislikes the genre `Horror'. Interestingly, when increasing the dimension from 2 to 3, we leave those two stereotypes approximately unchanged---we simply add a new stereotype; see figure~\ref{fig:ML_1M_genre_profiles_of_stereotypes} (right). The newly emergent stereotype has preferences for the genres `Action', `Adventure', `Children's', `Fantasy' and dislikes both Documentaries and movies from the Horror genre. All probabilities are large due to the \emph{selection bias towards high ratings}. Filling in missing entries with low ratings affects the size of those probabilities but leaves the structure of the towers approximately unchanged.

%


Figure~\ref{fig:ML_1M_hierarchy} serves as an example for how expressive computed tag-hierarchies are. The hierarchy from figure~\ref{fig:ML_1M_hierarchy} visualizes the pattern of `$\subseteq_{\varepsilon}$'-relations between movie genre for the value $\varepsilon=1/3$. To decrease the complexity of figure~\ref{fig:ML_1M_hierarchy}, we excluded the genres `Film-Noir' and `War' from the figure. Movies from these genres are rated highly by a majority of users and thus, all genres are connected to these genres. 


\begin{figure}[tbp]
\centering
\includegraphics[width=0.75\columnwidth]{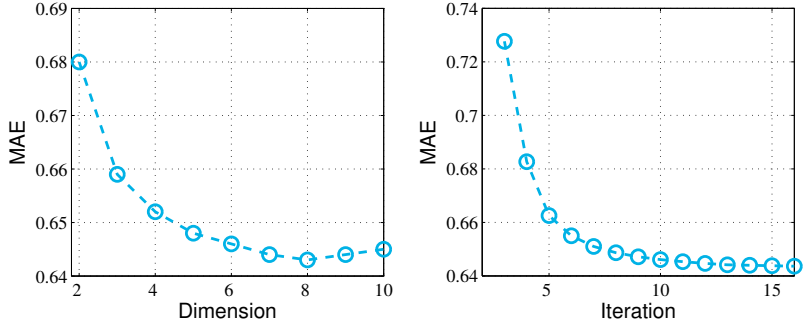}
\caption{Dependence on $D$ and iteration (when $D=8$).}
\label{fig:ML_1M_MAE_as_fn_of_dim}
\end{figure}

\emph{Evaluation of predictive power.} We evaluate mean-squared-error (MAE) and root-mean-squared-error (RMSE) through 5-fold cross validation with 0.8-to-0.2 data splitting. All results were computed using 16 iterations. Figure~\ref{fig:ML_1M_MAE_as_fn_of_dim} (left) shows how MAE depends on $D$, and Figure~\ref{fig:ML_1M_MAE_as_fn_of_dim} (right) shows that the proposed algorithm converges smoothly. In appendix~\ref{sect:MAE.vs.RMSE} we argue why MAE is less sensitive to outliers and therefore, we think that MAE should be preferred over RMSE. Table~\ref{Table.with.numerical.results} compares NNMs with popular recommender systems (see appendix~\ref{sect:configs}). We notice that Algorithm~\ref{Alg:Constrained.least.squares.for.NNM} outperforms SVD++~\cite{koren2008factorization} in MAE. Note that neither SVD++ nor NNMs were trained to minimize MAE. We evaluated the previously known methods by using the LibRec library~\cite{guolibrec}.

%
%

\begin{table}[htdp]
\caption{MAE and RMSE for MovieLens 100K and 1M. Table~\ref{Table:configs} in the appendix summarizes the configurations of all algorithms.}
\begin{center}
\begin{tabular}{|l|c|c|c|c|}
\hline
								& MAE (100K)	& RMSE (100K)	& MAE (1M)	& RMSE (1M)		 \\ 
\hline
UserKNN~\cite{resnick1994grouplens}	& 0.737		& 0.944		& 0.703		& 0.905		 		\\
ItemKNN~\cite{rendle2009bpr}			& 0.723		& 0.924		& 0.688		& 0.876		 	 		\\
NMF~\cite{lee2001algorithms}			& 0.747		& 0.949		& 0.727		& 0.920		 		\\
SVD++~\cite{koren2008factorization}	& 0.719		& {\bf 0.912 }	& 0.668		& {\bf 0.851}						\\
NNM								& {\bf 0.699}	& 0.984		& {\bf 0.643}	& 0.920		\\
\hline
\end{tabular}
\end{center}
\label{Table.with.numerical.results}
\end{table}%

\section{Related work}

In recommendation we are interested in the co-occurence of pairs $(u,i)$ of users $u$ and items $i$. Depending on the application, $(u,i)$ may be interpreted as \emph{`$u$ likes item $i$'}, \emph{`$u$ attends event $i$'}, etc. In \emph{aspect models}, \emph{pLSA}~\cite{hofmann1999latent,hofmann1999probabilistic} (and similarly in their extension \emph{LDA}~\cite{blei2003latent}) we model $(u,i)$ as random variable with distribution
\beq\label{fwfefeg}
	\mathbb{P}[u,i] = \sum_{k=1}^K \; \mathbb{P}[u|k] \; \mathbb{P}[i|k].
\eeq
Hence, $\mathbb{P}[u,i]$ is expressed as a inner product of two vectors: $(\mathbb{P}[u|k])_{k \in K}$ and $(\mathbb{P}[i|k])_{k \in K}$. This is reminiscent of~\eqref{few45jkljfnk} where we expressed the probability of the event \emph{`$u$ rates $i$ with $z$ stars'} in terms of the inner product between $\vec{E}_{iz}$ and $\vec{p}_u$. Therefore, one half of the inner product~\eqref{few45jkljfnk} agrees with the identity~\eqref{fwfefeg} as both $(\mathbb{P}[u|k])_{k \in K}$ and $\vec{p}_u$ are probability distributions. However, aspect models and NNMs disagree on the other half (i.e., $\vec{E}_{iz}$) because $\vec{E}_{iz}$ is constrained through the existence of $\vec{E}_{i1}, ..., \vec{E}_{iz-1}, \vec{E}_{iz+1}, .., \vec{E}_{iZ}$ such that $(\vec{E}_{i1}, ..., \vec{E}_{iZ}) \in \mathcal{M}$. More importantly however, the difference between aspect models and NNMs lies in the different interpretations of $(\mathbb{P}[u|k])_{k \in K}$ and $(\mathbb{P}[i|k])_{k \in K}$ on the one hand and $\vec{p}_u$ and $\vec{E}_{iz}$ on the other hand. 

Similarly, more general NMF-based models~\cite{ma2011recommender} agree with NNM-based models in that they make prediction in terms of inner products of nonnegative vectors but they differ from NNM-based models through different interpretations and regularizations of those nonnegative vectors.

These new interpretations of $\vec{p}_u$ and $\vec{E}_{iz}$ allow us to deal with situations where $z \in [Z]$ is a categorical random variable and we can extract hierarchical structures from NNMs in a straightforward manner. Aspect models can deal with both of these tasks too. However, we think that dealing with these tasks in terms of aspect models is less natural then dealing with these tasks with NNMs. For instance, to model multiple outcomes like ($z \in [5]$) we need to first decide on a particular graphical model (see section~2.3 in~\cite{hofmann1999latent}). Moreover, to extract hierarchical classifications of topics, we need to imagine generative processes like the nested Chinese restaurant process~\cite{blei2010nested}, or we need to employ nested hierarchical Dirichlet processes~\cite{paisley2015nested}. 

On the other hand, \emph{probabilistic matrix factorization} (PMF,~\cite{mnih2007probabilistic}) leads to another interesting and related class of models where we assume that entries $R_{ui}$ of the rating matrix are independent Gaussian random variables with mean $\vec{U}_u^T \vec{V}_i$ and variance $\sigma$. PMF-based models also describe ratings as samples of an underlying random variable and the distribution of that random variable is parameterize symmetrically in terms of vectors $\vec{U}_u$ assigned to users and $\vec{V}_i$ assigned to items. However, for PMF we need to assume that the random variables $R_{ui}$ are normally distributed. In NNM-based models, we do not need to assume anything about the distribution of $R_{ui}$ once the underlying dimension $D$ has been fixed by cross-validation. PMF would allow for the extraction of hierarchical orderings as discussed here but PMF does not allow for the interpretation of data through stereotypes because PMF corresponds to a specific infinite-dimensional NNM (PMF refers to continuous distributions).

The evaluation of NNMs in the extreme multi-label setting~\cite{agrawal2013multi,hsu2009multi,ji2008extracting,prabhu2014fastxml} is still outstanding.

\section{Conclusion}

We introduced NNMs at the example of item recommendation. We discussed in which way these models meet the criteria \emph{predictive power}, \emph{computational tractability} and \emph{interpretability} that are ideally met by recommender systems. The main strength of NNMs is their high level of interpretability. This quality can be used to characterize users' behavior in an \emph{interpretable} manner, and this quality can be used to derive hierarchical orderings of properties of items and users. Fortunately, as indicated by numerical experiments, these features of NNMs do not come at the price of sacrificing neither predictive power nor computational tractability. Hence, we believe that NNMs will prove valuable in recommendation and beyond.

\section{Acknowledgments}

I thank Patrick Pletscher and Sharon Wulff for interesting and fruitful discussions. I acknowledge funding by the ARO grant Contract Number W911NF-12-0486, and I acknowledge funding by the Swiss National Science Foundation through a postdoctoral fellowship.



\appendix

\section{MAE vs RMSE}\label{sect:MAE.vs.RMSE}


Assume the considered items are movies. Then, it can happen that a user $u$ likes movie $i$ but rates it badly because of the particular reason that one of the actors is a member of scientology. Hence, the reason for the poor rating is independent of user $u$'s movie taste; rather it is a consequence of user $u$ being atheist. It seems very unlikely that information of that kind is captured by the few numbers we use to describe a person's movie taste. Moreover, in practice, we would presumably not have enough data about movie $i$ to infer that one of the actors is a member of scientology. Hence, even extremely good models for capturing movie tastes will make a few predictions that are entirely wrong---nothing else should be expected. In conclusion, when measuring the quality of a model in terms of an error metric we may want to consider error metrics that are not too sensitive to a few outliers (i.e., predictions that are entirely wrong). RMSE is a scaled $l_2$-distance and MAE is a scaled $l_1$-distance. Hence, RMSE is much more sensitive to outliers than MAE. For that reason MAE appears to be more meaningful than RMSE. Moreover, there is an unavoidable tradeoff between enforcing low MAE and low RMSE---a recommender system cannot be near-optimal in both error measures.

\section{Configurations of algorithms}\label{sect:configs}

Table~\ref{Table:configs} specifies the configurations of algorithms as called by the Java LibRec library~\cite{guolibrec}. 

\begin{table}[htdp]
\caption{Configurations of SVD++~\cite{koren2008factorization}, NMF~\cite{lee2001algorithms}, ItemKNN~\cite{rendle2009bpr} and UserKNN as computed using the LibRec library~\cite{guolibrec}. These parameter choices are not necessarily optimal.}
 \centering
 \begin{tabular}{|c|l|l|}
 \hline
 & \multicolumn{1}{c|}{dataset} & \multicolumn{1}{c|}{configuration} \\
 \hline 
 \parbox[t]{2mm}{\multirow{3}{*}{\rotatebox[origin=c]{90}{NMF}}} 
 & ml-100K  	& num.factors=100, max.iter=10 \\[4pt]
 & ml-1M  		& num.factors=300, max.iter=10 \\[4pt]
  \hline
  \parbox[t]{2mm}{\multirow{3}{*}{\rotatebox[origin=c]{90}{ SVD++\hspace{28pt}}}} 
 & ml-100K 	& num.factors=5, max.iter=100, \\  && learn.rate=0.01 -max -1 -bold-driver,\\ && reg.lambda=0.1 -u 0.1 -i 0.1 -b 0.1 \\ && -s 0.001 \\[2pt]
 & ml-1M 		& num.factors=10, max.iter=80, \\  && learn.rate=0.005 -max -1 -bold-driver,\\ && reg.lambda=0.05 -u 0.05 -i 0.05 -b 0.05 \\ && -s 0.001 \\[2pt]
 \hline
 \parbox[t]{2mm}{\multirow{3}{*}{\rotatebox[origin=c]{90}{ItemKNN  }}} 
 & ml-100K  	& similarity=PCC, num.shrinkage=2500,\\ && num.neighbors=40 \\[4pt]
 & ml-1M  		& similarity=PCC, num.shrinkage=10, \\ && num.neighbors=80 \\[4pt]
 \hline
 \parbox[t]{2mm}{\multirow{3}{*}{\rotatebox[origin=c]{90}{UserKNN  }}} 
 & ml-100K  	& similarity=PCC, num.shrinkage=25, \\ && num.neighbors=60 \\[4pt]
 & ml-1M  		& similarity=PCC, num.shrinkage=25, \\ && num.neighbors=80\\[4pt]
 \hline
\parbox[t]{2mm}{\multirow{3}{*}{\rotatebox[origin=c]{90}{ NNM}}} 
 & ml-100K  	& $D=3$, max.iter $ = 16$ \\[4pt]
 & ml-1M  		& $D=8$, max.iter $ = 16$ \\[4pt]
 \hline
 \end{tabular}
 \label{Table:configs}
 \end{table}
 
  \begin{figure}[h]
\centering
\begin{subfigure}{.5\textwidth}
  \centering
  \includegraphics[width=1\linewidth]{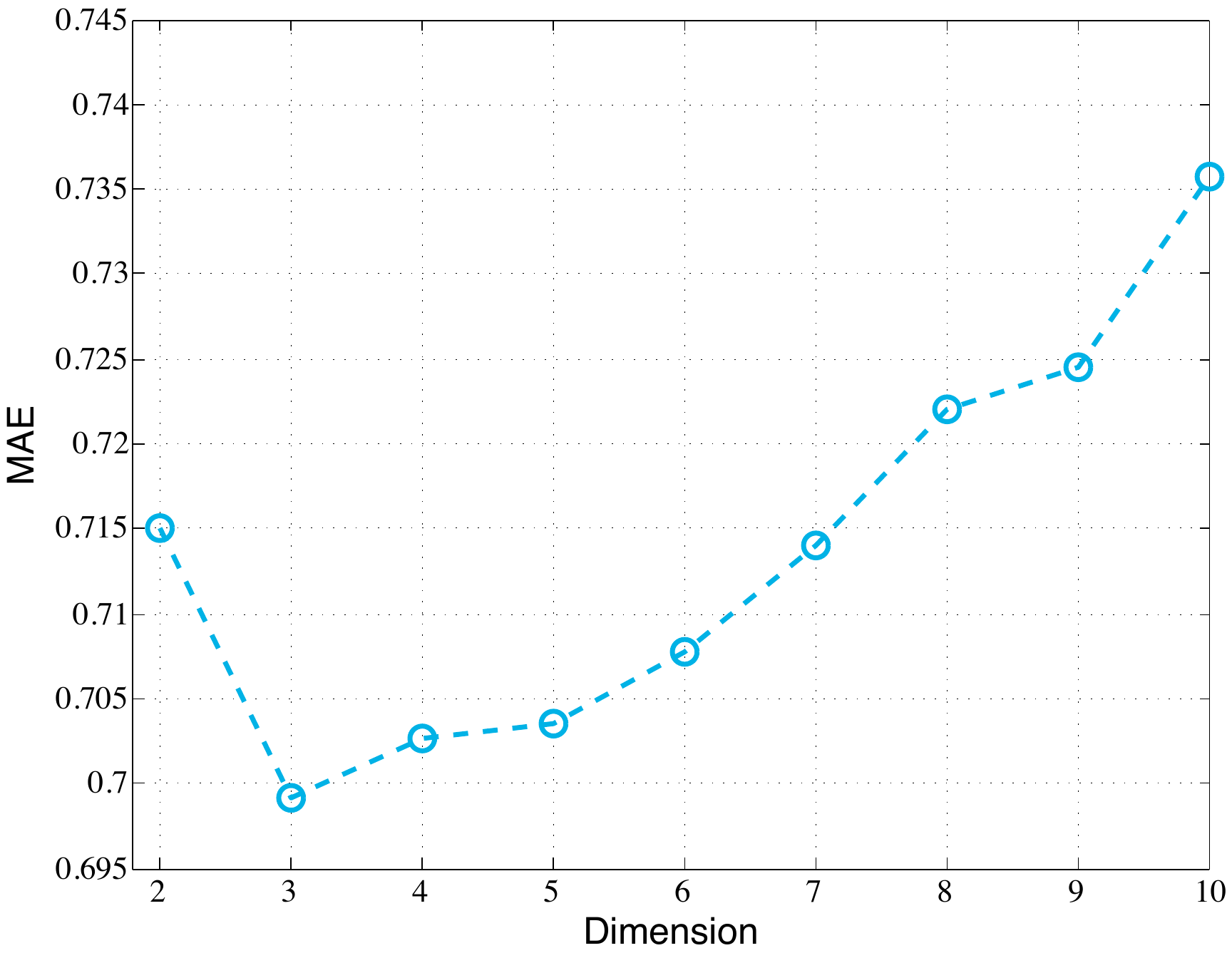}
\end{subfigure}%
\begin{subfigure}{.5\textwidth}
  \centering
  \includegraphics[width=1\linewidth]{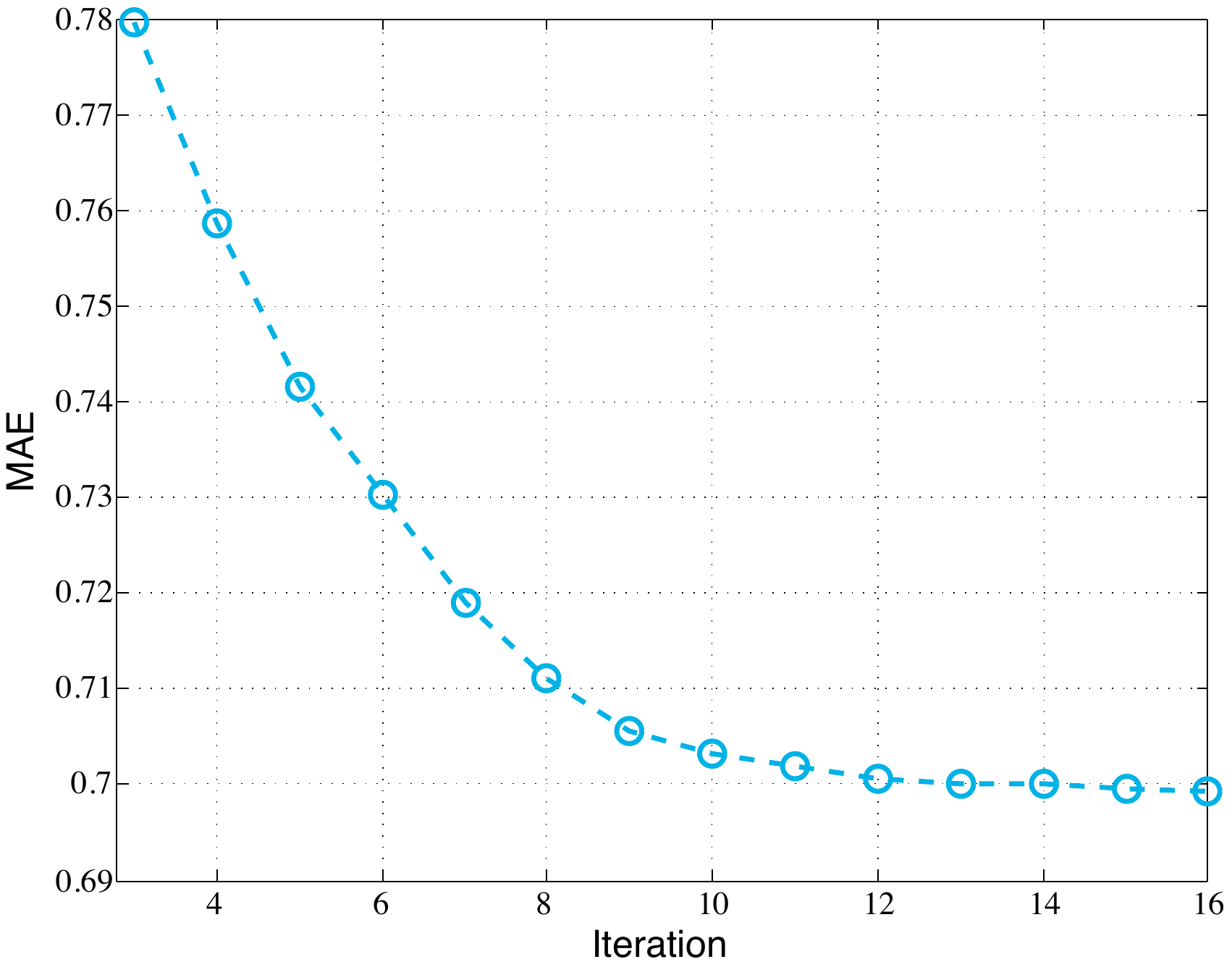}
\end{subfigure}
\caption{Left: Dependence on $D$; MovieLens 100K dataset. Right: Dependence on iteration; MovieLens 100K dataset.}
\label{fig:recall.as.fn.of.dim.MPM.and.recall.as.fn.of.iteration.MPM.100K}
\end{figure}

\end{document}